\documentclass[10pt,journal,compsoc]{IEEEtran}

\usepackage[T1]{fontenc}
\usepackage{graphicx}
\usepackage{amsmath,amssymb}
\usepackage{booktabs}
\usepackage{multirow}
\usepackage{array}
\usepackage{xspace}
\usepackage[nocompress]{cite}
\usepackage[hidelinks]{hyperref}
\usepackage{url}

\graphicspath{{assets/}}

\title{From Dense Prediction to Visual Editing:\\
Structured Supervision for Unified Image and Video Creation}
\author{Zhefan Rao$^{1*}$, Bin Zou$^{2*}$,
Xuanhua He$^1$, Chong Hou Choi$^2$,\\
Yanheng Li$^3$, Rui Liu$^{2\dagger}$, Haoxuan Che$^{1\dagger\ddagger}$, and Qifeng Chen$^{1\dagger}$\\[2pt]
{\small $^1$ The Hong Kong University of Science and Technology}\\
{\small $^2$ Celia Research HK}\\
{\small $^3$ City University of Hong Kong}%
\IEEEcompsocitemizethanks{%
\IEEEcompsocthanksitem Zhefan Rao, Xuanhua He, and Qifeng Chen are with The Hong Kong University of Science and Technology; Bin Zou, Haoxuan Che, Chong Hou Choi, and Rui Liu are with Celia Research HK; Yanheng Li is with City University of Hong Kong. Zhefan Rao and Bin Zou contributed equally to this work. Haoxuan Che, Rui Liu, and Qifeng Chen are the corresponding authors. Haoxuan Che is the project leader.}}

\begin{document}
\maketitle

\begin{abstract}
Unified image and video creation requires a model to follow diverse instructions while preserving identity, geometry, and temporal structure from visual context. However, semantic-only conditioning and creation-only training do not explicitly supervise the local structure needed for precise, temporally consistent editing. We therefore formulate depth and surface-normal prediction as image-form denoising targets, using these dense tasks as structured visual supervision within the same creation interface. Our framework decouples semantic interpretation from spatially aligned visual injection while sharing one multimodal diffusion transformer (MMDiT) backbone across all tasks. Mutual Context Attention (MCA), a paired-video data-construction procedure, and a progressive training curriculum then connect the learned structural cues to temporally localized editing and reference-conditioned creation. A single checkpoint obtains the highest overall score in the reported comparison of unified systems (4.15); adding dense supervision improves OpenVE Overall from 3.98 to 4.06 and Local Add from 3.92 to 4.18. These results support a deliberately bounded conclusion: perception-oriented dense supervision transfers useful structural knowledge to downstream creation, especially editing locality and preservation; we do not claim superiority as a standalone dense predictor.
\end{abstract}


\section{Introduction}
\label{sec:introduction}

Unified visual creation asks a single model to generate and edit both images and videos from language and visual context. Its scope includes text-to-image/video generation, subject-driven generation, instruction-based editing, reference-based generation and editing, and propagation from sparsely edited keyframes. Large video diffusion models have established strong synthesis priors~\cite{hong2022cogvideo,kong2024hunyuanvideo,hunyuanvideo2025,wan2025wan}, while recent unified systems broaden the supported condition types~\cite{jiang2025vace,ye2025unicunifiedincontextvideo,ju2025editverse,wei2025univideo,chen2026vino}. The remaining challenge is to make this breadth coexist with faithful control in one checkpoint.

Every task in this interface requires both semantic intent and preservation of spatial or temporal structure. A text instruction specifies \emph{what} to create or change, whereas a source image, video, subject, or keyframe specifies \emph{where}, \emph{when}, and relative to \emph{which} geometry the change should occur. Generation can tolerate many plausible spatial realizations; editing and reference-conditioned creation instead require the model to distinguish editable content from evidence that must remain fixed. The same requirement extends through time, because a locally correct frame can still produce an inconsistent video.

Lossy high-level conditioning makes these requirements fail in recognizable ways. Abstract visual tokens are effective for cross-modal semantics, but need not retain aligned evidence about object boundaries, occlusion, surface orientation, or motion continuity. Consequently, a model may obey an instruction while drifting from a reference identity, modifying unedited regions, distorting scene geometry, or flickering across frames. Existing editors mitigate individual symptoms with attention control or feature propagation~\cite{geyer2023tokenflow,qi2023fatezero,yang2025videograin,qin2024instructvid2vid}, yet the shared backbone is rarely supervised to predict the dense structure on which preservation depends.

Our central insight is to use depth and surface-normal prediction as structured visual supervision for creation. Expressed as image-form denoising targets, depth exposes relative layout, occlusion, and boundaries, while normals expose local surface geometry and orientation. These tasks can therefore train the same parameters and objective used for visual synthesis, without a perception-specific output head or an additional inference-time module. We hypothesize a specific form of transfer: learning to reconstruct dense structure improves structure-sensitive creation behavior, rather than necessarily producing a state-of-the-art dense predictor.

Our unified visual creation framework realizes this idea through a common condition package and two complementary conditioning pathways. Each example supplies an instruction, an optional primary visual input, optional auxiliary references, and a denoising target. A semantic pathway uses frozen vision-language features to interpret instructions and cross-modal relations, while a visual pathway injects spatially aligned VAE latents directly into a shared multimodal diffusion transformer (MMDiT). This separation retains high-level intent without forcing fine visual evidence through a lossy semantic bottleneck.

The learning path combines structured supervision, editing-pair construction, and curriculum design as parts of one system. Depth and normal targets directly teach spatial organization; Mutual Context Attention (MCA) constructs aligned source--target videos whose edits can begin at different times; and progressive training moves from generation and reconstruction, through instruction editing and dense prediction, to reference-conditioned creation. Thus, dense tasks teach the backbone what scene structure is, while MCA editing pairs teach where and when that structure should be preserved or changed.

The evidence consistently supports this downstream-transfer account across complementary evaluations. One checkpoint covers seven task types and reaches 4.15 overall in the reported unified comparison. With the architecture and editing mixture fixed, dense supervision raises OpenVE Overall from 3.98 to 4.06 and Local Add from 3.92 to 4.18. The editing instantiation obtains 4.43 on OpenVE-Bench and 4.61 on InsEdit-Bench under the manuscript protocol, while a controlled data study improves the downstream score from 3.96 to 4.08 by replacing half of 4K conventional pairs with 2K MCA pairs. Together, these results localize the benefit to task coverage, structure-sensitive editing, and more informative paired supervision.

\begin{figure*}[t]
    \centering
    \includegraphics[width=\textwidth]{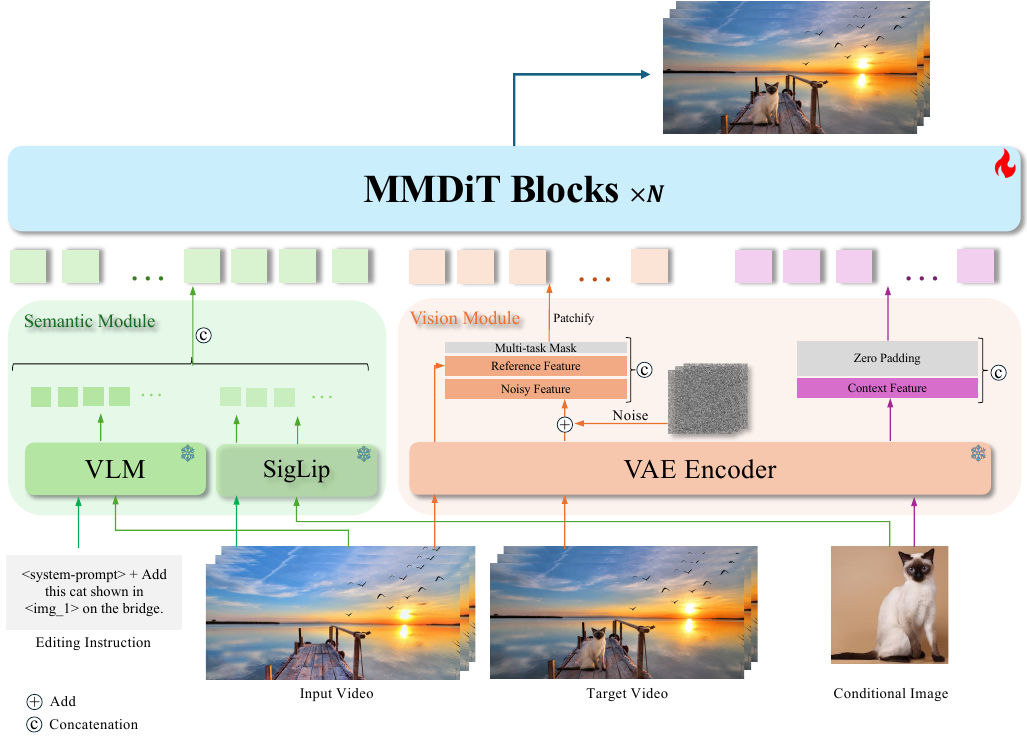}
    \caption{\textbf{Unified visual creation framework.} The semantic pathway encodes instruction intent and cross-modal relations, while the visual pathway preserves spatial evidence from sources, references, and controls. Noisy target tokens, clean condition tokens, and their role identifiers are processed by one shared MMDiT. Creation outputs and depth/normal structured-supervision targets use the same target slots and denoising objective.}
    \label{fig:unified_framework}
\end{figure*}

Our contributions are threefold:
\begin{itemize}
    \item We introduce dense prediction as structured visual supervision for a shared image/video creation backbone, using depth and surface-normal targets to improve downstream structural preservation without adding inference-time components.
    \item We develop a unified denoising framework that decouples semantic intent from spatially aligned visual evidence and supports seven generation and editing task types in one checkpoint.
    \item We establish a structure-aware training recipe that combines dense supervision, MCA-generated editing pairs, and progressive training, supported by unified and editing-focused experiments with controlled ablations.
\end{itemize}

\section{Related Work}
\label{sec:related_work}

\subsection{Unified Image and Video Creation}
Large diffusion models learn strong priors for appearance and motion from scalable image and video pretraining~\cite{hong2022cogvideo,kong2024hunyuanvideo,hunyuanvideo2025,wan2025wan}. Building on these priors, VACE~\cite{jiang2025vace}, UNIC~\cite{ye2025unicunifiedincontextvideo}, EditVerse~\cite{ju2025editverse}, UniVideo~\cite{wei2025univideo}, Kling-Omni~\cite{klingteam2025klingomnitechnicalreport}, and VINO~\cite{chen2026vino} expand a shared model toward generation, editing, and mixed image/video context. This progress demonstrates that task breadth can be represented by one interface. Our focus is the complementary learning question: how the shared backbone can be explicitly supervised for the local geometry needed to preserve sources and references across that interface.

\subsection{Instruction- and Reference-Based Editing}
Instruction-based image editing established language as a flexible transformation interface, from InstructPix2Pix~\cite{Brooks2023InstructPix2Pix} to recent work on data quality, rewards, and benchmark coverage~\cite{nhredit,gpt-image-edit,liu2025step1x-edit,luo2025editscore,ye2025imgedit}. Video methods additionally impose temporal constraints through attention manipulation, feature propagation, or learned spatiotemporal control~\cite{qi2023fatezero,geyer2023tokenflow,cong2023flatten,kara2024rave,ku2024anyv2v,yang2025videograin}. Direct instruction editors such as InsViE, ICVE, Ditto, and OpenVE scale paired data and adaptation~\cite{Wu2025InsVie,Liao2025ICVE,Bai2025Ditto,openve}. These methods primarily address text-specified transformations; our framework places instruction editing within a broader shared task space and targets structural preservation through both aligned visual conditioning and dense supervision.

Reference-based methods transfer identity, appearance, or style from exemplars. DreamBooth~\cite{ruiz2023dreambooth} adapts a model to a subject, while IP-Adapter~\cite{ye2023ipadapter} injects image features through dedicated cross-attention. In-context systems generalize this principle to heterogeneous source and reference inputs~\cite{mou2025instructx,ye2025unicunifiedincontextvideo,ju2025editverse}. Unlike approaches that compress every visual input into semantic tokens, we retain a separate path for spatially aligned VAE features; unlike task-specific adapters, that path and the denoising backbone are shared across image/video generation and editing.

\subsection{Generation and Visual Perception}
Generative pretraining can yield representations useful for dense perception~\cite{xu2024diffusion}, and recent work reframes depth, normals, segmentation, and related outputs as image generation within a generalist model~\cite{gabeur2026imagegenerators}. That line studies transfer from image generation toward visual perception. We study the reverse, complementary direction: perception-oriented depth and normal supervision is placed in the creation training mixture, and its value is measured by transfer back to generation and editing behavior. Accordingly, our empirical claim concerns improved downstream editing locality and structural preservation, not general perceptual intelligence or dense-prediction state of the art.

\subsection{Structured Supervision and Editing-Data Construction}
Dense maps have long provided explicit generation controls. ControlNet~\cite{zhang2023adding} and T2I-Adapter~\cite{mou2024t2i}, for example, consume edges, depth, or other spatial maps to constrain a generated output. Our formulation differs in direction and use: the model \emph{predicts} depth and normals as training targets, but requires no dense control at creation time. This converts dense structure from an external condition into structured visual supervision for the shared denoising parameters.

Editing data quality is equally important, particularly for video. Large paired datasets commonly edit an anchor frame and propagate the result through the sequence~\cite{Wu2025InsVie,Bai2025Ditto,openve}; this is scalable, but it underrepresents edits that begin after the first frame or last for only part of a clip. Mutual self-attention control has shown that corresponding content can be coupled across related generations~\cite{cao2023masactrltuningfreemutualselfattention}. MCA adapts this principle into a paired-video construction mechanism: source and target clips share context during synthesis while an edit mask can vary temporally, producing aligned supervision for localized changes. Dense targets and MCA pairs therefore play distinct roles---the former exposes scene structure, whereas the latter specifies where and when that structure should change.

\begin{table*}[t]
    \centering
    \caption{\textbf{Unified task formulation.} Diverse creation and dense-prediction tasks vary the contents of a shared condition package rather than the model interface.}
    \label{tab:task_taxonomy}
    \resizebox{\textwidth}{!}{%
    \begin{tabular}{llll}
        \toprule
        Task family & Primary input & Optional context & Target behavior \\
        \midrule
        Text-to-image/video generation & Text prompt & None & Synthesize an image or video from scratch \\
        In-context generation & Text prompt & Subject, frame, sketch, layout, pose, depth, normal & Generate content consistent with visual context \\
        Instruction-based editing & Source image/video + instruction & Mask or structured control & Apply requested edits while preserving untouched regions \\
        In-context editing & Source image/video & Reference image/video, edited keyframes & Transfer style, identity, structure, or edits across time \\
        Dense prediction & Source image/video + instruction & Target representation type & Predict depth or normal maps as image-form outputs \\
        \bottomrule
    \end{tabular}%
    }
\end{table*}

\begin{figure*}[t]
    \centering
    \includegraphics[width=\textwidth]{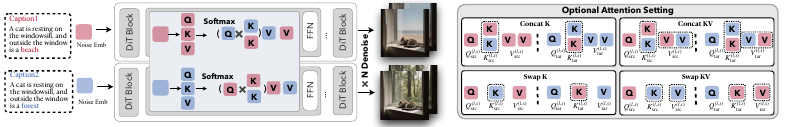}
    \caption{\textbf{Mutual Context Attention for aligned pair synthesis.} Source and target branches are denoised in one shared DiT; scheduled cross-branch key/value interaction aligns scene layout, identity, and motion while branch-specific queries retain the requested edit.}
    \label{fig:mca_attention}
\end{figure*}

\section{Unified Visual Creation Framework}
\label{sec:framework}

\begin{figure*}[t]
    \centering
    \includegraphics[width=0.98\textwidth]{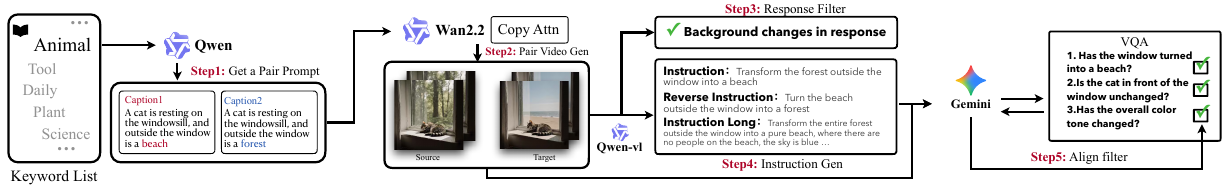}
    \caption{\textbf{Automatic construction and filtering of video editing pairs.} Prompt expansion is followed by MCA-coupled source/target synthesis, instruction generation, response filtering, and multi-round VQA verification.}
    \label{fig:mca_data_pipeline}
\end{figure*}

\begin{figure}[t]
    \centering
    \includegraphics[width=\columnwidth]{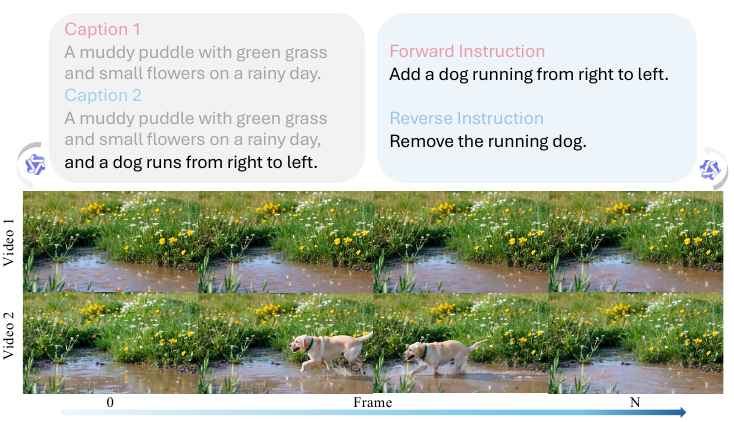}
    \caption{\textbf{MCA-generated editing pairs.} Examples include edits whose onset occurs after the first frame or only within an intermediate interval, while unchanged content remains aligned.}
    \label{fig:mca_data_examples}
\end{figure}

Our framework casts image and video generation, editing, reference-conditioned creation, and dense prediction as conditional denoising in one latent space. Figure~\ref{fig:unified_framework} gives the common computation graph: semantic tokens explain the requested operation, clean visual tokens retain aligned evidence, and noisy target tokens are predicted by a shared MMDiT. Task identity is expressed by the supplied conditions and token roles rather than by task-specific networks.

\subsection{Unified Task Formulation}
\label{sec:task_formulation}

We represent the condition of each training or inference instance as
\begin{equation}
    \mathcal{C}=(p,x_{\mathrm{src}},\mathcal{R},\rho),
    \label{eq:condition_package}
\end{equation}
where $p$ is a generation prompt or editing instruction, $x_{\mathrm{src}}$ is an optional primary source image or video, $\mathcal{R}=\{r_k\}_{k=1}^{K}$ is an optional set of references or structured controls, and $\rho$ contains learned token-role identifiers. The desired output is $y$, represented as an image or video even when it encodes a dense prediction. Roles distinguish source, reference/control, target, and padding slots after latent patchification. Hence identical visual values can be interpreted differently depending on whether they must be preserved, used as auxiliary evidence, or denoised as the output.

Tasks alter the contents of $\mathcal{C}$ and the target $y$, but not the model interface. For text-guided generation, only $p$ is present and $y$ is synthesized from noise. Instruction editing supplies $x_{\mathrm{src}}$ and a transformation in $p$; reference-conditioned tasks additionally populate $\mathcal{R}$ with subjects, styles, layouts, or edited keyframes. For dense prediction, $x_{\mathrm{src}}$ is the observed image or video, $p$ specifies the representation to predict, and $y$ is the corresponding depth or normal map. Table~\ref{tab:task_taxonomy} summarizes the five task families and their seven image/video instantiations evaluated in this work.

\subsection{Decoupled Semantic and Visual Conditioning}
\label{sec:conditioning}

The semantic pathway determines intent and cross-modal binding. A frozen Qwen2.5-VL encoder receives $p$ together with available source/reference thumbnails and interprets relations such as which referenced subject should be inserted or which source attribute should change. A frozen SigLIP encoder supplies compact visual semantic features that stabilize identity, category, and style cues, while Glyph-ByT5 supplies features specialized for text rendering. Lightweight learned projectors map the three feature streams to a common hidden dimension and concatenate them into semantic condition tokens $S(\mathcal{C})$. This pathway is therefore responsible for instruction semantics and reference assignment, rather than pixel-accurate reconstruction.

The visual pathway preserves spatially aligned evidence. A frozen VAE maps $x_{\mathrm{src}}$, every $r_k$, and the training target $y$ into a common latent representation. Source and reference/control latents remain clean, are patchified with the same spatiotemporal layout as the target, and receive the corresponding role embeddings from $\rho$. Image examples use the identical layout with a temporal extent of one frame. Because clean visual tokens share the output coordinate system, attention can directly recover local appearance, boundaries, camera framing, and framewise correspondence.

This decoupling avoids assigning incompatible responsibilities to one bottleneck. Semantic intent need not carry pixel-aligned geometry, and spatial evidence is not compressed solely into a small set of abstract tokens. The MMDiT instead combines both forms of evidence at every block: semantic tokens answer what relation is requested, while source and reference latents retain what must be copied, preserved, or spatially transformed.

\subsection{Shared MMDiT Denoising}
\label{sec:mmdit}

Let $z_y$ be the frozen-VAE latent of target $y$. At a sampled diffusion timestep $t$, we form
\begin{equation}
    z_t=\alpha_t z_y+\sigma_t\epsilon,
    \qquad \epsilon\sim\mathcal{N}(0,I),
    \label{eq:forward_noise}
\end{equation}
where $\alpha_t$ and $\sigma_t$ follow the backbone noise schedule. The target sequence contains patchified $z_t$ with target roles; clean source/reference latents, semantic tokens, and the timestep embedding jointly condition MMDiT attention. The shared training objective is
\begin{equation}
    \mathcal{L}_{\mathrm{denoise}}=
    \mathbb{E}_{t,\epsilon}
    \left[\left\lVert
    \epsilon-\epsilon_\theta(z_t,\mathcal{C},t)
    \right\rVert_2^2\right].
    \label{eq:denoising_loss}
\end{equation}
The loss is evaluated only on target slots; clean source, reference/control, and semantic tokens act exclusively as conditions. Joint attention nevertheless allows each target location to retrieve high-level intent and fine visual evidence in the same operation.

At inference, target slots begin as Gaussian noise and are iteratively denoised while condition slots remain fixed. Switching from a generated video to an edited image therefore changes only $\mathcal{C}$, $\rho$, and the temporal extent, not the backbone, objective, or decoder. We initialize MMDiT from a pretrained video generator~\cite{hunyuanvideo2025} to retain appearance and motion priors; frozen semantic encoders and the VAE provide stable input spaces while the shared denoising parameters learn the task mixture described next.

\begin{table*}[t]
    \centering
    \caption{\textbf{Unified image/video creation.} Qwen3-VL-32B scores use a 0--5 scale for one unified checkpoint per method; Overall aggregates T2I, T2V, S2V, TI2I, TV2V, II2I, and IV2V.}
    \label{tab:unified_results}
    \resizebox{\textwidth}{!}{%
    \begin{tabular}{lcccccccc}
        \toprule
        Method & Unified Overall $\uparrow$ & \multicolumn{3}{c}{Generation} & \multicolumn{4}{c}{Editing} \\
        \cmidrule(lr){3-5}\cmidrule(lr){6-9}
        & & Overall $\uparrow$ & Quality $\uparrow$ & Semantic $\uparrow$ & Overall $\uparrow$ & Quality $\uparrow$ & Semantic $\uparrow$ & IFS $\uparrow$ \\
        \midrule
        UniVideo~\cite{wei2025univideo} & 4.03 & 4.08 & 4.12 & 4.04 & 3.98 & 4.01 & 3.95 & 3.90 \\
        VINO~\cite{chen2026vino} & 4.12 & \textbf{4.16} & \textbf{4.19} & 4.13 & 4.08 & 4.11 & 4.06 & 4.03 \\
        \textbf{Ours} & \textbf{4.15} & 4.15 & 4.17 & 4.13 & \textbf{4.16} & \textbf{4.18} & \textbf{4.15} & \textbf{4.14} \\
        \bottomrule
    \end{tabular}%
    }
\end{table*}

\begin{figure*}[t]
    \centering
    \includegraphics[width=\textwidth]{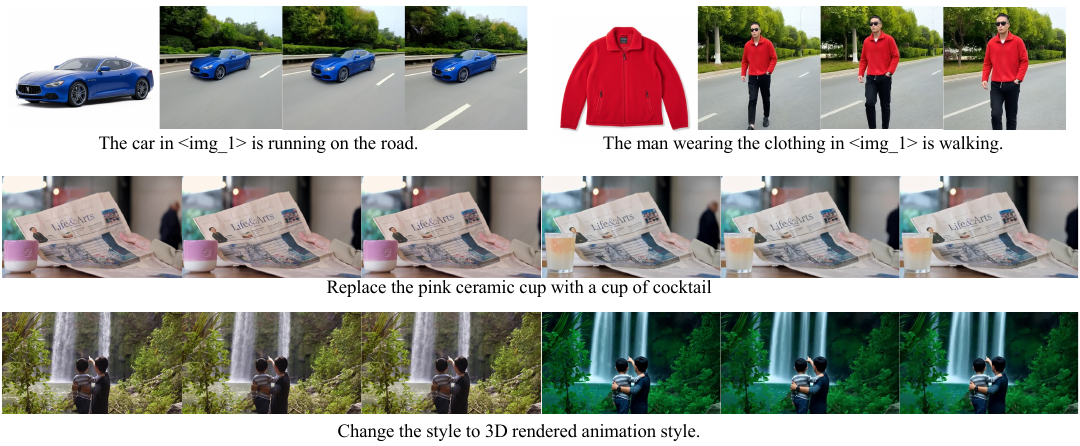}
    \caption{\textbf{Unified creation examples.} One checkpoint performs subject-to-video generation, text-instruction video editing, and image-reference video editing while retaining reference identity, unedited scene content, and temporal structure.}
    \label{fig:unified_results}
\end{figure*}

\section{Learning Structural Understanding for Creation}
\label{sec:structured_learning}

The shared interface in Section~\ref{sec:framework} makes structural learning a data and objective problem: any image-form target can supervise the same MMDiT parameters used for creation. We exploit this property in two complementary ways. Dense prediction supplies direct geometric targets, while aligned editing pairs expose controlled changes within otherwise corresponding videos. A progressive curriculum integrates both sources without changing the inference graph.

\subsection{Dense Prediction as Structured Visual Supervision}
\label{sec:dense_supervision}

Precise editing depends on relations that ordinary creation pairs do not explicitly label. Locality requires object boundaries and depth ordering to separate an edit target from its surroundings; reference preservation additionally depends on scene layout, occlusion, and surface geometry. A reconstruction loss can in principle learn these cues implicitly, but it does not isolate them from appearance, texture, or many equally plausible generations. We instead make the relevant structure an explicit prediction target during shared training.

Specifically, normalized depth is rendered as an RGB image, and surface normals are represented as three-channel orientation maps. The frozen VAE encodes either representation into $z_y$, while the observed image or video occupies $x_{\mathrm{src}}$ and an instruction such as ``convert this video to a depth map'' specifies the task. These examples then use exactly the condition package in Eq.~\eqref{eq:condition_package} and denoising loss in Eq.~\eqref{eq:denoising_loss}. Depth emphasizes global layout, relative distance, and occlusion; normals complement it with local orientation and fine boundary geometry.

This formulation provides structured visual supervision to the same attention and feed-forward parameters that perform generation and editing. It adds neither a perception-specific head nor an inference-time structural module. Dense examples are simply sampled into the creation mixture, and no depth or normal map is required when the trained checkpoint edits or generates content. The intended advantage is therefore transfer to structure-sensitive creation; standalone dense-prediction accuracy is outside the evaluation claim of this work.

\subsection{MCA-Based Paired-Video Construction}
\label{sec:mca_data}

Dense targets describe scene structure, but an editor also needs paired evidence of a permitted change. Existing scalable pipelines often edit the first frame and propagate that result, which favors changes active for the entire clip and may introduce artifacts around the anchor. We construct more diverse pairs with Mutual Context Attention (MCA), a controlled interaction between two video-denoising branches that preserves shared context while allowing the target branch to realize a specified edit.

For branch $b\in\{\mathrm{src},\mathrm{tar}\}$, let $\neg b$ denote the other branch and let $Q_b$, $K_b$, and $V_b$ be attention states at a selected DiT layer and denoising step. MCA keeps queries branch-specific and selects the key/value context used by
\begin{equation}
    \operatorname{Attn}_b=
    \operatorname{softmax}\!\left(
    \frac{Q_b\bar K_b^{\mathsf T}}{\sqrt d}
    \right)\bar V_b.
    \label{eq:mca_attention}
\end{equation}
Four atomic interactions span different coupling strengths. Because paired branches have equal token counts, their contexts are dimensionally defined as
\begin{equation}
\begin{aligned}
\textsc{Concat K}:~&(\bar K_b,\bar V_b)=([K_b;K_{\neg b}],[V_b;V_b]),\\
\textsc{Concat KV}:~&(\bar K_b,\bar V_b)=([K_b;K_{\neg b}],[V_b;V_{\neg b}]),\\
\textsc{Swap K}:~&(\bar K_b,\bar V_b)=(K_{\neg b},V_b),\\
\textsc{Swap KV}:~&(\bar K_b,\bar V_b)=(K_{\neg b},V_{\neg b}),
\end{aligned}
\label{eq:mca_policies}
\end{equation}
where $[\cdot;\cdot]$ concatenates along the token dimension and \textsc{Concat K} repeats branch-specific values to match the expanded key sequence. Concat variants permit softer correspondence, whereas swap variants more strongly lock coarse layout and motion. MCA is activated only at selected layers and steps: strong early coupling aligns scene structure, softer middle coupling retains editability, and late release avoids copying texture artifacts.

The task-specific policies instantiate this shared schedule without introducing separate mechanisms:
\begin{itemize}
    \item \textbf{Object insertion and removal.} One branch retains self-attention, while the other uses \textsc{Swap KV} in early denoising, \textsc{Concat KV} in the middle, and self-attention late. The asymmetric release permits a large semantic difference without moving unchanged context.
    \item \textbf{Local object modification.} Both branches use early \textsc{Swap KV} and middle \textsc{Concat KV}, first locking coarse structure and then allowing the edited region to deviate.
    \item \textbf{Background replacement.} \textsc{Swap K} is emphasized in shallow-to-middle layers during early and middle steps, anchoring contextual geometry while retaining foreground editability.
    \item \textbf{Color and material modification.} Early \textsc{Swap KV}/\textsc{Concat KV} protects instance geometry; middle \textsc{Concat K} then permits appearance changes while stabilizing shape and extent.
    \item \textbf{Motion and viewpoint transformation.} \textsc{Concat KV} is used over the selected layers and steps, softly sharing identity and scene context while allowing different motion states or camera relations.
\end{itemize}
These category-level policies use the same paired denoising graph and attention operators throughout; only the layer/time policy changes.

Figure~\ref{fig:mca_data_pipeline} shows the automatic pipeline around this interaction. An LLM expands sampled keywords into related source/target prompts; a shared DiT denoises both clips jointly with task-aware MCA; invalid responses are filtered; a VLM converts each retained pair into editing instructions; and multi-round VQA verification rejects pairs with poor alignment, instruction mismatch, or visual defects. All retained videos are standardized to 480p, 16 frames per second, and three seconds. This construction separates pair alignment from any anchor-frame editor or propagation model, so data quality can improve with the underlying video generator.

The resulting InsEdit Data contains approximately 300K source--target--instruction triples across object insertion/removal, attribute and material changes, background replacement, motion transformation, and viewpoint change. Because the two branches are generated jointly, the target may share an unedited prefix or suffix with the source; an edit can begin after the first frame, end before the last frame, or occupy only an intermediate interval. Figure~\ref{fig:mca_data_examples} illustrates these non-first-frame onsets.

Dense supervision and aligned pairs thus address different parts of the learning problem. Dense targets teach which structural relations exist in a scene. Aligned editing pairs teach which relations should remain stable and which may change under an instruction. Temporally localized MCA pairs additionally teach when a change should begin or end, connecting spatial preservation to the temporal behavior required by video editing.

\subsection{Unified Progressive Curriculum}
\label{sec:curriculum}

\begin{table}[t]
    \centering
    \caption{\textbf{Progressive training curriculum.} The model interface and loss remain fixed; only task sampling changes.}
    \label{tab:training_curriculum}
    \resizebox{\columnwidth}{!}{%
    \begin{tabular}{p{0.10\columnwidth}p{0.42\columnwidth}p{0.36\columnwidth}}
        \toprule
        Stage & Objectives & Capability introduced \\
        \midrule
        I & Generation; VLM reconstruction; consistency preservation (7:2:1) & Condition alignment and prior retention \\
        II & Instruction editing; depth/normals; MCA pairs; generation replay & Local editing and explicit structural supervision \\
        III & Subject/reference generation; reference editing; keyframe propagation; multi-reference composition & Reference binding and compositional creation \\
        \bottomrule
    \end{tabular}%
    }
\end{table}

Introducing every task at once can disrupt the pretrained synthesis prior before the new condition pathways have acquired stable meanings. We therefore retain one architecture and loss while progressively changing only the sampled tasks. This ordering first aligns conditions, then learns structure-aware editing, and finally increases reference composition complexity.

\textbf{Stage I: conditioning alignment.} Text-guided generation, VLM-guided reconstruction, and consistency preservation are sampled in a 7:2:1 mixture. Generation replay retains the inherited visual prior; reconstruction teaches the projected semantic tokens to specify content; and consistency examples teach clean visual slots to preserve identity and layout. The loss remains restricted to noisy target slots.

\textbf{Stage II: editing and structured supervision.} Instruction editing is mixed with depth/normal targets, MCA-generated pairs, and a reduced amount of generation replay. Image and video examples use a default 4:1 sampling ratio: images provide broad and spatially dense editing supervision, whereas videos introduce the temporal correspondences emphasized by MCA. Training both target types concurrently lets the backbone associate editing locality with explicit scene structure rather than learning the two capabilities in isolated fine-tuning stages.

\textbf{Stage III: reference-conditioned creation.} The mixture expands to subject/reference generation, reference-based editing, edited-keyframe propagation, and multi-reference composition. Dense supervision is retained at a reduced regularization ratio, without altering the denoising objective. Thus, the final checkpoint preserves the common condition package while learning increasingly complex bindings among instructions, sources, references, and targets.

\begin{figure*}[t]
    \centering
    \begin{minipage}[t]{0.48\textwidth}
        \centering
        \includegraphics[width=\linewidth]{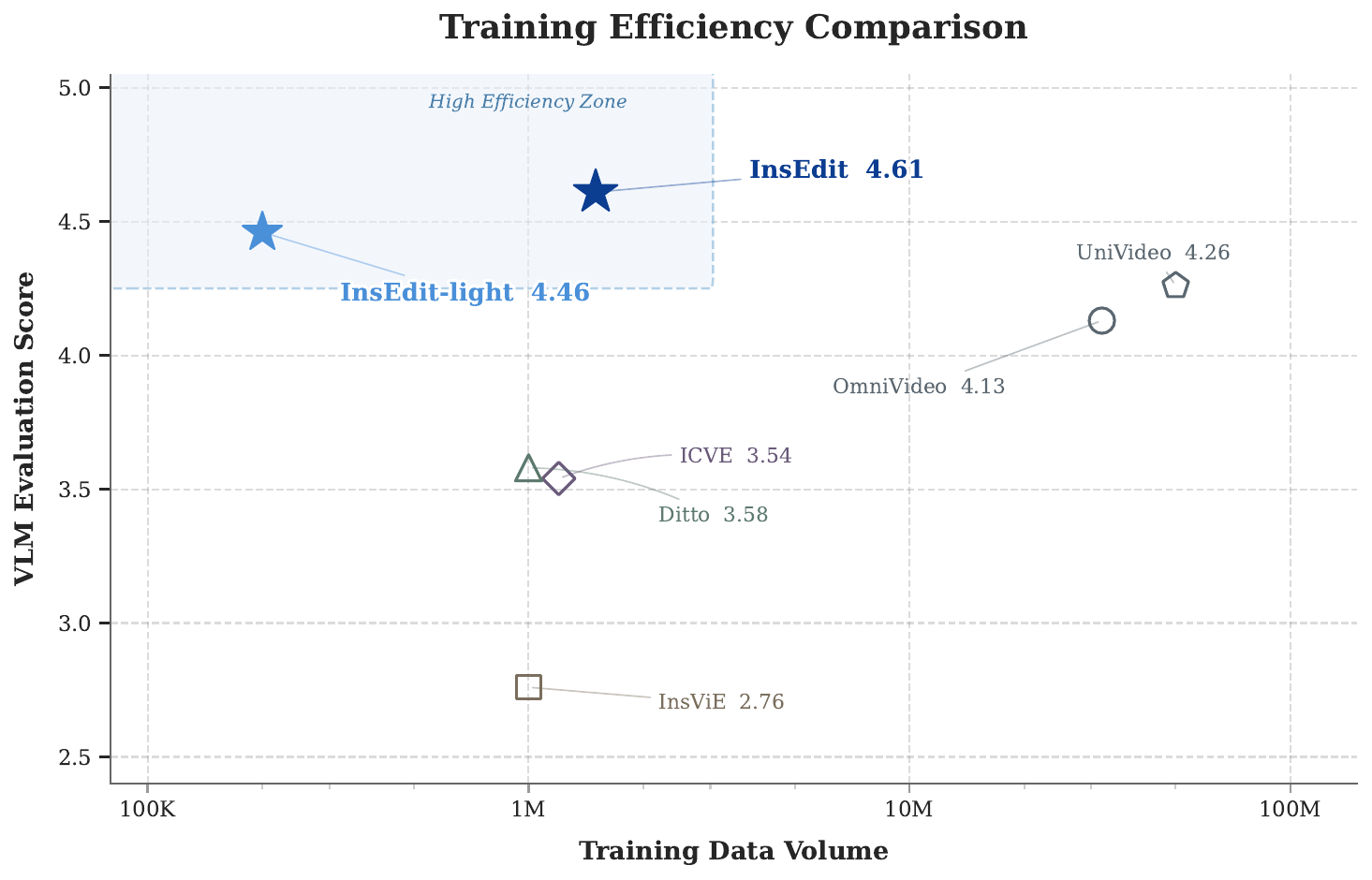}\\[-1mm]
        \small (a) Editing performance versus video-editing sample scale.
    \end{minipage}\hfill
    \begin{minipage}[t]{0.48\textwidth}
        \centering
        \includegraphics[width=\linewidth]{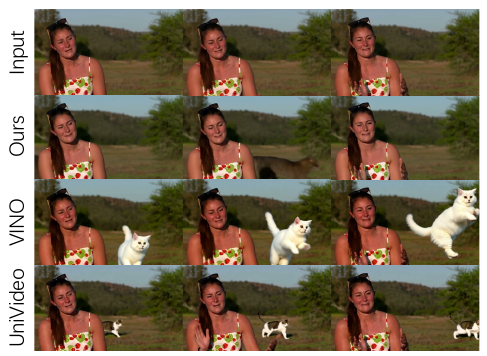}\\[-1mm]
        \small (b) An edit beginning after an unchanged prefix.
    \end{minipage}
    \caption{\textbf{MCA data value and adaptation behavior.} (a) The editing instantiation reaches strong benchmark performance with $O(100)$K video editing samples while also using larger image-editing and generation sources. (b) Temporally localized editing demonstrates a behavior underrepresented by first-frame propagation data.}
    \label{fig:training_efficiency}
    \label{fig:temporally_localized_edit}
\end{figure*}

\section{Experiments}
\label{sec:experiments}

\begin{table*}[t]
    \centering
    \caption{\textbf{Downstream transfer from dense supervision on OpenVE-Bench.} Scores evaluate instruction video editing, not standalone depth/normal accuracy.}
    \label{tab:dense_transfer}
    \resizebox{0.9\textwidth}{!}{%
    \begin{tabular}{lcccccc}
        \toprule
        Variant & Overall $\uparrow$ & Local Add & Local Remove & Local Change & Background Change & Global Style \\
        \midrule
        w/o Dense Prediction & 3.98 & 3.92 & 3.84 & 3.89 & 3.76 & 3.89 \\
        w/ Dense Prediction & \textbf{4.06} & \textbf{4.18} & \textbf{4.02} & \textbf{4.11} & \textbf{3.95} & \textbf{4.06} \\
        \bottomrule
    \end{tabular}%
    }
\end{table*}

\begin{table*}[t]
    \centering
    \caption{\textbf{Instruction video editing.} OpenVE-Bench reports representative categories; Table~\ref{tab:openve_full} gives the complete category breakdown. Latency is minutes per 81-frame 480p output.}
    \label{tab:video_editing}
    \resizebox{\textwidth}{!}{%
    \begin{tabular}{lccccccccccc}
        \toprule
        \multirow{2}{*}{Method} & \multicolumn{6}{c}{OpenVE-Bench} & \multicolumn{4}{c}{InsEdit-Bench} & \multirow{2}{*}{Latency $\downarrow$} \\
        \cmidrule(lr){2-7}\cmidrule(lr){8-11}
        & Overall $\uparrow$ & Local Add & Local Remove & Local Change & Subtitle Edit & Creative Edit & Overall $\uparrow$ & IC $\uparrow$ & TVQ $\uparrow$ & URP $\uparrow$ & \\
        \midrule
        VACE-14B~\cite{jiang2025vace} & 3.01 & 1.76 & 3.99 & 2.47 & 4.41 & 2.17 & 3.08 & 2.39 & 2.64 & 4.21 & 9.08 \\
        OmniVideo~\cite{tan2025omni} & 3.66 & 2.80 & 4.52 & 3.75 & \textbf{4.95} & 1.13 & 4.13 & 4.04 & 4.04 & 4.32 & 20.58 \\
        InsViE~\cite{Wu2025InsVie} & 3.25 & 2.25 & 3.56 & 2.82 & 4.77 & 3.36 & 2.76 & 2.19 & 2.29 & 3.80 & 1.06 \\
        Lucy-Edit~\cite{decart2025lucyedit} & 3.77 & 3.92 & 3.95 & 3.93 & 4.23 & 4.19 & 3.64 & 3.24 & 3.38 & 4.30 & \textbf{0.60} \\
        ICVE~\cite{Liao2025ICVE} & 3.76 & 3.77 & 4.50 & 3.87 & 4.68 & 3.54 & 3.54 & 2.91 & 2.94 & 4.79 & 20.60 \\
        Ditto~\cite{Bai2025Ditto} & 3.44 & 2.48 & 3.53 & 2.89 & 3.69 & 4.14 & 3.58 & 3.45 & 3.52 & 3.77 & 7.33 \\
        UniVideo~\cite{wei2025univideo} & 4.21 & 4.41 & 4.46 & 4.33 & 4.56 & 4.33 & 4.26 & 4.11 & 4.07 & 4.59 & 21.50 \\
        VINO~\cite{chen2026vino} & 4.34 & 4.43 & 4.45 & 4.41 & 3.39 & 4.60 & 4.42 & 4.31 & 4.31 & 4.62 & 6.50 \\
        \midrule
        InsEdit (ours) & \textbf{4.43} & \textbf{4.78} & \textbf{4.64} & \textbf{4.71} & 4.73 & \textbf{4.66} & \textbf{4.61} & \textbf{4.50} & \textbf{4.54} & \textbf{4.80} & 1.95 \\
        \bottomrule
    \end{tabular}%
    }
\end{table*}

We evaluate the framework at three levels: breadth across unified creation tasks, transfer from structured visual supervision to editing, and the editing/data-design choices that support this transfer. Unless explicitly marked as an ablation, each comparison uses one checkpoint per method and the protocol associated with the named benchmark.

\subsection{Implementation Details}
We initialize the backbone from HunyuanVideo-1.5~\cite{hunyuanvideo2025}. Qwen2.5-VL, SigLIP, Glyph-ByT5, and the VAE remain frozen; MMDiT and the condition projectors are trainable. Optimization uses AdamW with learning rate $2\times10^{-5}$ and DeepSpeed ZeRO-2. Videos are trained at 480p and images at 720p. Inference uses 50 sampling steps without classifier-free guidance.

The training mixture contains distinct image and video sources. Image editing data is at the scale of $O(1)$M and supplies broad spatial transformations, while video editing data is at the scale of $O(100)$K and supplies temporal supervision; the latter includes retained samples from the MCA pipeline. Generation replay is also used as specified by the curriculum. Thus, $O(100)$K characterizes video \emph{editing} samples, not the total amount of image, video, generation, or pretraining data.

\subsection{Evaluation Protocol}
The unified evaluation covers seven task types: text-to-image (T2I), text-to-video (T2V), subject-to-video (S2V), text-instruction image editing (TI2I), text-instruction video editing (TV2V), image-reference image editing (II2I), and image-reference video editing (IV2V). T2I/T2V use 100 diverse prompts, S2V uses a subset of OpenS2V-Eval~\cite{yuan2025opens2vnexus}, TI2I uses GEdit~\cite{liu2025step1x-edit}, and TV2V uses OpenVE-Bench~\cite{openve}; fixed internally constructed reference-editing subsets exercise II2I/IV2V. A Qwen3-VL-32B judge reports overall, quality, and semantic scores on the unified protocol's 0--5 scale, with instruction-following score (IFS) additionally measuring whether an edit realizes the request.

For focused video-editing evaluation, we use OpenVE-Bench and InsEdit-Bench. InsEdit-Bench contains 82 cases selected from 80 Pexels source videos. Its 13 category counts are addition 15, removal 10, replacement 11, recoloring 6, retexturing 5, relocation 5, rescaling 7, background switching 8, weather switching 4, time switching 2, season switching 2, stylization 5, and relighting 2. The prompt split contains 47 short instructions, 25 long instructions, and 10 multi-instruction cases. Source clips were manually selected so that each requested change is meaningful and visually assessable.

For each method, the Qwen3-VL-32B-Instruct judge receives the source video, editing instruction, and generated output under common preprocessing, and assigns scores on a 1--5 scale. Instruction Compliance (IC) scores requested changes; Temporal Visual Quality (TVQ) scores framewise appearance and consistency through time; and Unedited Region Preservation (URP) scores retention outside the requested edit. Overall is the benchmark's aggregate assessment. We report the scoring dimensions and judge inputs, but do not reconstruct an unavailable verbatim judge prompt. Latency is measured on the same single GPU with FlashAttention-2~\cite{dao2023flashattention2} for 81 frames at 480p.

\subsection{Unified Creation Performance}
Does one checkpoint retain competitive generation while supporting all seven task types? Table~\ref{tab:unified_results} shows that our framework reaches the highest reported unified overall score, 4.15 versus 4.12 for VINO and 4.03 for UniVideo. Its generation score (4.15) is within 0.01 of VINO, while its editing overall score rises to 4.16 from 4.08. The comparison therefore supports breadth without an observed aggregate generation penalty, with the clearest advantage on editing.

\subsection{Does Structured Supervision Improve Editing?}
The central controlled question is whether adding depth/normal targets improves downstream creation when the editing setting is otherwise fixed. Table~\ref{tab:dense_transfer} answers this on OpenVE-Bench: structured visual supervision raises Overall from 3.98 to 4.06. The largest absolute gains occur for Local Add (3.92 to 4.18) and Local Change (3.89 to 4.11), while Local Remove, Background Change, and Global Style improve by 0.18, 0.19, and 0.17, respectively. This pattern is consistent with better use of boundaries, layout, and preserved context. It establishes downstream editing transfer under this training comparison; it does not measure standalone depth/normal quality or isolate extra data volume from target semantics.

\subsection{Instruction-Based Image and Video Editing}
How strong is the editing instantiation under dedicated benchmarks? Table~\ref{tab:video_editing} reports that InsEdit reaches 4.43 Overall on OpenVE-Bench and 4.61 on InsEdit-Bench, the highest reported scores among the compared open-source systems under this protocol. On InsEdit-Bench it also obtains 4.50 IC, 4.54 TVQ, and 4.80 URP. Its 1.95-minute latency is substantially below several high-scoring unified baselines, although Lucy-Edit and InsViE are faster; the quality result is therefore not presented as a latency optimum.

Table~\ref{tab:openve_full} expands the OpenVE-Bench comparison to all reported categories. InsEdit obtains the highest Overall, Local Add, Local Remove, Local Change, and Creative Edit scores among these rows, but does not lead Background Change, Global Style, Subtitle Edit, or Camera Edit. The complete breakdown therefore supports strong local editing without implying uniform superiority across every edit type.

\begin{table*}[t]
    \centering
    \caption{\textbf{Complete category-level OpenVE-Bench results.}}
    \label{tab:openve_full}
    \resizebox{\textwidth}{!}{%
    \begin{tabular}{lccccccccc}
        \toprule
        Method & Overall $\uparrow$ & Local Add & Local Remove & Local Change & Background Change & Global Style & Subtitle Edit & Creative Edit & Camera Edit \\
        \midrule
        VACE-14B~\cite{jiang2025vace} & 3.01 & 1.76 & 3.99 & 2.47 & 2.81 & 3.46 & 4.41 & 2.17 & 3.09 \\
        OmniVideo~\cite{tan2025omni} & 3.66 & 2.80 & 4.52 & 3.75 & 4.11 & 3.41 & 4.95 & 1.13 & 3.62 \\
        InsViE~\cite{Wu2025InsVie} & 3.25 & 2.25 & 3.56 & 2.82 & 2.68 & 3.63 & 4.77 & 3.36 & 3.61 \\
        Lucy-Edit~\cite{decart2025lucyedit} & 3.77 & 3.92 & 3.95 & 3.93 & 3.25 & 3.64 & 4.23 & 4.19 & 3.54 \\
        ICVE~\cite{Liao2025ICVE} & 3.76 & 3.77 & 4.50 & 3.87 & 3.51 & 3.87 & 4.68 & 3.54 & 2.84 \\
        Ditto~\cite{Bai2025Ditto} & 3.44 & 2.48 & 3.53 & 2.89 & 3.52 & 4.48 & 3.69 & 4.14 & 3.33 \\
        OpenVE-Edit~\cite{openve} & 3.89 & 3.41 & 3.50 & 3.80 & 4.10 & 4.24 & 3.98 & 3.71 & 3.25 \\
        VINO~\cite{chen2026vino} & 4.34 & 4.43 & 4.45 & 4.41 & 4.46 & 4.78 & 3.39 & 4.60 & 4.08 \\
        \midrule
        InsEdit (ours) & 4.43 & 4.78 & 4.64 & 4.71 & 3.99 & 4.20 & 4.73 & 4.66 & 3.62 \\
        \bottomrule
    \end{tabular}%
    }
\end{table*}

Treating images as single-frame videos also yields useful image editing, but the evidence is more limited. On GEdit (Table~\ref{tab:image_editing}), InsEdit obtains G\_SC/G\_PQ/G\_O of 6.98/7.77/6.72. G\_PQ is the best value among the reported open-source rows, whereas semantic consistency and overall score remain below Step1X-Edit and VINO. This is a secondary capability arising from joint image/video training rather than a claim of best overall image editing.

\begin{table}[t]
    \centering
    \caption{\textbf{Image editing on GEdit.} Bold marks the best value among reported open-source methods.}
    \label{tab:image_editing}
    \resizebox{0.95\columnwidth}{!}{%
    \begin{tabular}{lccc}
        \toprule
        Method & G\_SC $\uparrow$ & G\_PQ $\uparrow$ & G\_O $\uparrow$ \\
        \midrule
        Gemini2.5 & 7.48 & 8.30 & 7.17 \\
        GPT4o & 8.06 & 7.80 & 7.48 \\
        Seedream4 & 8.33 & 8.00 & 7.72 \\
        \midrule
        UniWorld-V1~\cite{lin2025uniworld} & 5.04 & 7.56 & 4.98 \\
        OmniGen2~\cite{wu2025omnigen2} & 6.79 & 6.68 & 6.18 \\
        Flux-Kontext-Dev~\cite{batifol2025fluxkontext} & 7.23 & 7.28 & 6.53 \\
        Bagel~\cite{deng2025bagel} & 7.52 & 6.69 & 6.54 \\
        Step1X-EditV1.1~\cite{liu2025step1x-edit} & \textbf{7.60} & 7.29 & 6.87 \\
        VINO~\cite{chen2026vino} & 7.26 & 7.71 & \textbf{6.88} \\
        \midrule
        InsEdit (ours) & 6.98 & \textbf{7.77} & 6.72 \\
        \bottomrule
    \end{tabular}%
    }
\end{table}

\begin{figure*}[t]
    \centering
    \includegraphics[width=\textwidth]{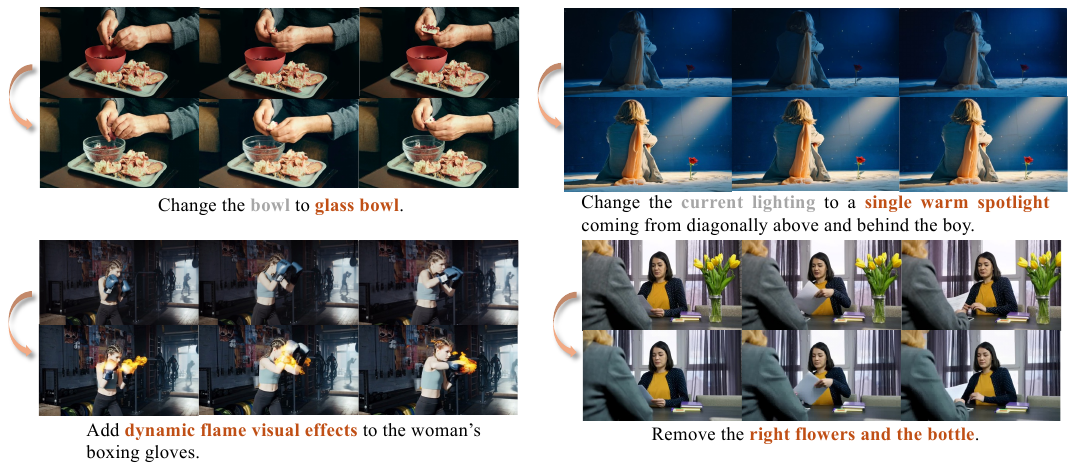}
    \caption{\textbf{Instruction video editing results.} Examples span local object, attribute, background, motion, and style changes while preserving content outside the instruction.}
    \label{fig:video_edit_results}
\end{figure*}

\begin{figure}[t]
    \centering
    \includegraphics[width=\columnwidth]{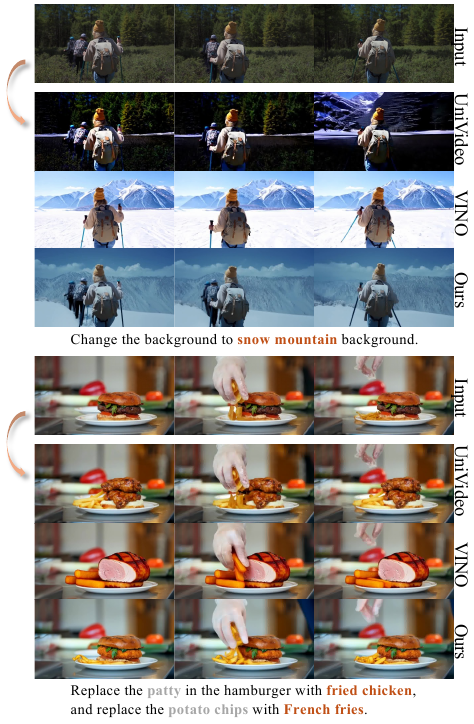}
    \caption{\textbf{Qualitative comparison.} InsEdit is compared with unified baselines on instruction compliance, temporal quality, and preservation of unedited regions.}
    \label{fig:video_edit_comparison}
\end{figure}

\subsection{MCA Data Quality and Data Efficiency}
Does MCA improve pair quality rather than merely increasing sample count? Under matched generator, prompt, seed, resolution, denoising, and filtering controls, Table~\ref{tab:mca_audit} gives MCA the highest source--target alignment (0.84), prompt fidelity (0.82), and temporal consistency (0.81), together with the lowest artifact rate (9.8\%). Alignment scores correspondence of nominally unchanged content, prompt fidelity scores the intended source--target difference, temporal consistency scores continuity within each clip, and artifact rate is the fraction failing visual-defect checks. These audit values compare favorably with independent generation (0.67/0.79/0.72, 18.6\%) and first-frame propagation (0.78/0.71/0.76, 15.4\%).

\begin{table}[t]
    \centering
    \caption{\textbf{Controlled audit of paired-video construction.} Generator, prompts, seeds, resolution, denoising budget, and filtering thresholds are fixed. Artifact is a rate ($\downarrow$); other metrics are scores ($\uparrow$).}
    \label{tab:mca_audit}
    \resizebox{\columnwidth}{!}{%
    \begin{tabular}{lcccc}
        \toprule
        Construction & Alignment & Prompt & Temporal & Artifact (\%) \\
        \midrule
        Independent generation & 0.67 & 0.79 & 0.72 & 18.6 \\
        First-frame propagation & 0.78 & 0.71 & 0.76 & 15.4 \\
        MCA-generated & \textbf{0.84} & \textbf{0.82} & \textbf{0.81} & \textbf{9.8} \\
        \bottomrule
    \end{tabular}%
    }
\end{table}

Table~\ref{tab:mca_category_audit} further reports an audited construction batch, rather than the entire approximately 300K corpus, across five grouped edit categories. Keep rates range from 78.5\% to 81.2\%; the temporal-edit group contains 33.7K raw and 26.9K retained temporally localized pairs.

\begin{table}[t]
    \centering
    \caption{\textbf{Category-level MCA data audit.} Raw/retained counts in the audited construction batch, keep rates.}
    \label{tab:mca_category_audit}
    \begin{tabular}{lrrc}
        \toprule
        Category & Raw & Kept & Keep rate \\
        \midrule
        Object add/remove & 38.4K & 30.6K & 79.7\% \\
        Attribute/action & 31.2K & 24.5K & 78.5\%  \\
        Background/weather & 26.8K & 21.4K & 79.9\%  \\
        Style/relighting & 18.6K & 15.1K & 81.2\%  \\
        Temporal edits & 33.7K & 26.9K & 79.8\% \\
        \bottomrule
    \end{tabular}
\end{table}

Across all rejected samples in this audit, the observed cause distribution is prompt not fully satisfied (24.8\%), source--target misalignment (20.6\%), temporal flicker or motion artifact (18.7\%), over-shared context suppressing the edit (15.2\%), and an over-edited source or weak preservation (20.7\%). These categories sum to 100\% and expose both alignment and editability failures of automatic construction.

The matched-budget downstream test in Table~\ref{tab:mca_downstream} replaces half of 4K conventional pairs with 2K MCA pairs while keeping the total at 4K. The InsEdit-Bench Overall score increases from 3.96 to 4.08. This result isolates a benefit within the stated small comparison, while the complete model still trains with $O(100)$K video editing and $O(1)$M image editing samples.

\begin{table}[t]
    \centering
    \caption{\textbf{Matched-budget downstream MCA study.} Both variants use 4K video editing pairs; only pair composition changes.}
    \label{tab:mca_downstream}
    \begin{tabular}{lc}
        \toprule
        Video editing data & InsEdit-Bench Overall $\uparrow$ \\
        \midrule
        4K non-MCA & 3.96 \\
        2K non-MCA + 2K MCA & \textbf{4.08} \\
        \bottomrule
    \end{tabular}
\end{table}

\subsection{Training and Architecture Ablations}
Which adaptation choices matter after the shared framework is fixed? Table~\ref{tab:ablation_modeling} shows that adding generation replay with single-frame SigLIP raises Overall from 4.41 to 4.46. Using three SigLIP frames reduces the score to 4.30--4.32, and removing visual input from the VLM produces the largest decline to 4.10. The latter result supports source-aware semantic conditioning, while the former indicates that more semantic frames are not automatically beneficial.

\begin{table}[t]
    \centering
    \caption{\textbf{Stage-II modeling variants on InsEdit-Bench.} Generation replay, VLM visual input, and SigLIP temporal input are varied.}
    \label{tab:ablation_modeling}
    \resizebox{\columnwidth}{!}{%
    \begin{tabular}{lcccc}
        \toprule
        Variant & Overall $\uparrow$ & IC $\uparrow$ & TVQ $\uparrow$ & URP $\uparrow$ \\
        \midrule
        Edit-only Baseline & 4.41 & 4.24 & 4.30 & 4.70 \\
        + Gen Data w/ SigLIP-1 & \textbf{4.46} & \textbf{4.32} & \textbf{4.33} & \textbf{4.72} \\
        + Gen Data w/ SigLIP-3 & 4.30 & 4.09 & 4.17 & 4.63 \\
        + Gen Data w/ SigLIP-3 Avg & 4.32 & 4.11 & 4.21 & 4.65 \\
        + Gen Data w/o VLM Vision & 4.10 & 3.83 & 3.80 & 4.52 \\
        \bottomrule
    \end{tabular}%
    }
\end{table}

Table~\ref{tab:ablation_recipe} examines the edit-only branch. Removing all Stage-I initialization lowers Overall from 4.41 to 4.15; removing consistency or VLM-reconstruction initialization gives intermediate scores of 4.26 and 4.29. Relative to the default 4:1 image/video mixture, 1:1 and 1:4 reach 4.34 and 4.27. Training on only short or only long instructions also underperforms the default three-way prompt mixture, and a short/long mixture recovers only to 4.21. These results motivate the chosen initialization, image-heavy sampling, and prompt diversity, without attributing the central dense-transfer gain to these separate recipe studies.

\begin{table}[t]
    \centering
    \caption{\textbf{Edit-only Stage-II recipe on InsEdit-Bench.} Groups vary Stage-I initialization, image/video ratio, and prompt mixture.}
    \label{tab:ablation_recipe}
    \resizebox{\columnwidth}{!}{%
    \begin{tabular}{clcccc}
        \toprule
        Group & Variant & Overall $\uparrow$ & IC $\uparrow$ & TVQ $\uparrow$ & URP $\uparrow$ \\
        \midrule
        \textit{Baseline} & Edit-only Baseline & \textbf{4.41} & \textbf{4.24} & \textbf{4.30} & \textbf{4.70} \\
        \midrule
        \multirow{3}{*}{\textit{Initialization}} & w/o Stage 1 Init & 4.15 & 3.94 & 3.95 & 4.57 \\
        & w/o Stage 1 Consistency Init & 4.26 & 3.99 & 4.04 & 4.74 \\
        & w/o Stage 1 VLM-Recon Init & 4.29 & 4.09 & 4.15 & 4.63 \\
        \midrule
        \multirow{2}{*}{\textit{Data Ratio}} & Image:Video = 1:1 & 4.34 & 4.15 & 4.16 & 4.71 \\
        & Image:Video = 1:4 & 4.27 & 4.07 & 4.12 & 4.62 \\
        \midrule
        \multirow{3}{*}{\textit{Prompt Format}} & Short Instruction only & 4.14 & 3.87 & 4.00 & 4.55 \\
        & Long Instruction only & 4.19 & 3.90 & 3.99 & 4.67 \\
        & Short/Long Mix & 4.21 & 3.88 & 4.02 & 4.72 \\
        \bottomrule
    \end{tabular}%
    }
\end{table}

\subsection{Qualitative Analysis}
The qualitative results expose both task breadth and the preservation behaviors measured above. Figure~\ref{fig:unified_results} spans subject-driven generation, instruction editing, and reference editing. The evidence for MCA then appears at three complementary levels: Fig.~\ref{fig:mca_data_examples} shows constructed source--target pairs, Fig.~\ref{fig:temporally_localized_edit}(b) isolates an edit whose onset is not the first frame, and Figs.~\ref{fig:video_edit_results} and~\ref{fig:video_edit_comparison} show downstream outputs with surrounding appearance and motion retained. Together, these examples connect the constructed supervision to the quantitatively evaluated behavior without replacing the pair audit or benchmark tables. Remaining failures include long-video identity drift, imprecise changes to small objects, and conflicting bindings when multiple references specify similar content; the Discussion examines these boundaries.

\section{Discussion and Limitations}
\label{sec:discussion}

The results suggest a bidirectional relationship between visual creation and perception-oriented training. Generative pretraining supplies broad priors for appearance, motion, and language-conditioned synthesis, while depth and normal targets make geometry-oriented structure explicit in the same image-form output space. Sharing the denoising parameters allows these signals to meet without adding a task-specific perception head.

The downstream pattern is consistent with this structural account but should not be read more broadly. The largest dense-supervision gains occur for local addition/change and background-sensitive editing, where boundaries, depth ordering, and preservation are particularly relevant. This alignment between mechanism and behavior strengthens the transfer hypothesis, but it is neither a standalone perception benchmark nor evidence of general visual understanding.

Dense supervision and MCA pairs are complementary rather than interchangeable. Dense targets encode which spatial relations exist in a scene; aligned pairs encode which relations may change and which should remain stable under an instruction. Temporally localized pairs further expose when a permitted change begins or ends. Their roles therefore connect structural representation, edit locality, and temporal control at different points in the same training process.

\subsection{Limitations}
The present evidence has two important control limitations. First, we do not report independent depth or normal benchmarks, so the results cannot establish dense-prediction competitiveness. Second, the dense ablation lacks an equal-size generic-data control; it shows that the tested dense-supervised mixture improves editing, but cannot separate target semantics from every effect of additional data. Evaluation also depends substantially on a VLM judge. InsEdit-Bench contains only 82 cases, and its category-level conclusions should therefore be confirmed on larger benchmarks and with human evaluation.

The operating regime introduces additional practical limits. Training and the principal video evaluation use 480p outputs, leaving higher-resolution locality untested. Qualitative inspection also reveals identity drift in longer videos, insufficient precision for small objects, and binding conflicts when multiple references specify similar subjects or attributes. These behaviors delimit the current checkpoint even when short-clip aggregate scores are strong.

\section{Conclusion}

Unified image and video creation requires more than broad instruction following: editing and reference-conditioned synthesis must preserve local geometry and temporal structure. We investigated depth and surface-normal prediction as structured visual supervision, expressed through the same image-form denoising interface as creation tasks.

The resulting framework combines a shared MMDiT with decoupled semantic and spatial pathways, MCA-generated paired videos, and progressive multi-task training. The reported evidence covers seven task types, yields the highest unified overall score among the compared unified baselines, improves OpenVE editing metrics when dense supervision is added, and shows stronger MCA pair audits and matched-budget downstream performance. These findings support transfer from perception-oriented supervision to structure-sensitive creation.

Our conclusion is limited to that downstream transfer; standalone dense-task superiority and general visual-understanding improvements are not established. Future work should test optical flow and segmentation as additional structured targets, extend training and evaluation to longer videos, add human evaluation, and measure depth/normal performance directly. Improving small-object precision and multi-reference binding is likewise necessary before the framework can support more demanding production settings.

\bibliographystyle{IEEEtran}
\bibliography{references}
\end{document}